\documentclass[conference]{IEEEtran}

\usepackage{cite}
\usepackage{amsmath,amssymb,amsfonts}
\usepackage{algorithmic}
\usepackage{graphicx}
\usepackage{textcomp}
\usepackage{xcolor}

\usepackage{multirow}
\usepackage{booktabs}

\IEEEoverridecommandlockouts
\def\BibTeX{{\rm B\kern-.05em{\sc i\kern-.025em b}\kern-.08em
    T\kern-.1667em\lower.7ex\hbox{E}\kern-.125emX}}
\begin{document}

\title{Cross-Lingual Transfer for Natural Language Inference via Multilingual Prompt Translator}


\author{\IEEEauthorblockN{Xiaoyu Qiu}
\IEEEauthorblockA{\textit{University of Science and Technology} \\
\textit{of China}\\
Hefei, China \\
qiuxy@mail.ustc.edu.cn}
\and
\IEEEauthorblockN{Yuechen Wang}
\IEEEauthorblockA{\textit{University of Science and Technology} \\
\textit{of China}\\
Hefei, China \\
wyc9725@mail.ustc.edu.cn}
\and
\IEEEauthorblockN{Jiaxin Shi}
\IEEEauthorblockA{\textit{Cloud BU, Huawei Technologies} \\
Beijing, China \\
shijiaxin3@huawei.com}
\and
\IEEEauthorblockN{Wengang Zhou}
\IEEEauthorblockA{\textit{University of Science and Technology of China}\\
Hefei, China \\
zhwg@ustc.edu.cn}
\and
\IEEEauthorblockN{Houqiang Li}
\IEEEauthorblockA{\textit{University of Science and Technology of China}\\
Hefei, China \\
lihq@ustc.edu.cn}
}

\maketitle

\begin{abstract}
Based on multilingual pre-trained models, cross-lingual transfer with prompt learning has shown promising effectiveness, where soft prompt learned in a source language is transferred to target languages for downstream tasks, particularly in the low-resource scenario. To efficiently transfer soft prompt, we propose a novel framework, Multilingual Prompt Translator (MPT), where a multilingual prompt translator is introduced to properly process crucial knowledge embedded in prompt by changing language knowledge while retaining task knowledge. Concretely, we first train prompt in source language and employ translator to translate it into target prompt. Besides, we extend an external corpus as auxiliary data, on which an alignment task for predicted answer probability is designed to convert language knowledge, thereby equipping target prompt with multilingual knowledge. In few-shot settings on XNLI, MPT demonstrates superiority over baselines by remarkable improvements. MPT is more prominent compared with vanilla prompting when transferring to languages quite distinct from source language.
\end{abstract}


\section{Introduction}
Recently, significant progress has been made on various NLP tasks by Pre-trained Language Models (PLMs)~\cite{devlin2018bert,brown2020language}. However, fine-tuning PLMs for downstream tasks often relies on tremendous manually annotated data, which is easily obtained for widely-used languages (e.g., English) but very hard and expensive for low-resource languages (e.g., Swahili). To this end, an abundance of works perform cross-lingual transfer~\cite{conneau2019cross,conneau2019unsupervised} based on multilingual PLMs by fine-tuning PLMs on a higher-resource language and then applying it in lower-resource languages. Notably, Conneau et al.~\cite{conneau2019unsupervised} achieve large performance gains in Cross-Lingual Understanding (XLU) for two low-resource languages, thus making up for the unfairness caused by the gap in different language resources.
\begin{figure}[t]
 \centering
 \includegraphics[width=0.9\columnwidth]{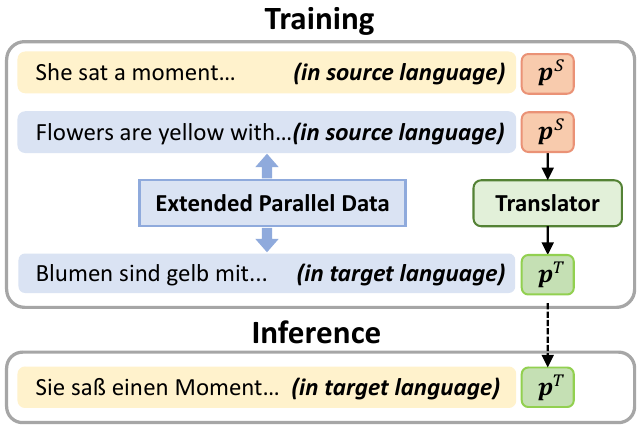}
 \caption{The source language training data in yellow and extended parallel data in blue are concatenated with the corresponding prompt for joint training. A multilingual prompt translator is employed to translate the source language prompt ($\boldsymbol{p}^S$) into multilingual prompt as target prompt ($\boldsymbol{p}^T$). In the inference, prediction is made on target language test data in yellow with translated $\boldsymbol{p}^T$.}
 \label{MPT0}
\end{figure}

Considering the notable superiority of prompt learning mentioned earlier, numerous studies ~\cite{zhao2021discrete,li2022learning,fu2022polyglot} have explored cross-lingual transfer approaches utilizing prompting. Zhao et al.~\cite{zhao2021discrete} leverage soft prompting by applying source language prompt to all languages in XLU. Nevertheless, Qi et al.~\cite{qi2022enhancing} indicate this approach fails to adequately capture language-specific characteristics. Therefore, they adopt cross-lingual hard prompt and enforce consistency in answer probability across languages via the Kullback-Leibler Divergence (KLD) loss, remarkably enhancing transferability. Yet, Liu et al.~\cite{liu2021gpt} shed light on that manually designed hard prompt is labor-intensive and performance is volatile since a single-word’s change can cause a drastic difference. 

 To address the above issues, we present Multilingual Prompt Translator (MPT), a novel method for enhancing cross-lingual transferability by translating soft prompt learned in source language into target prompt. As depicted in Figure~\ref{MPT0}, MPT employs a multi-task strategy for joint optimization during training. On one hand, we conduct the original task on source language training data to enable prompt to learn task knowledge. This involves processing a source language training sentence combined with source language prompt $\boldsymbol{p}^S$ through a PLM, resulting in a probability distribution for the predicted category. The optimization objective is to minimize CE loss between predicted category and gold label. On the other hand, we simultaneously execute a cross-lingual alignment task on an external parallel corpus to optimize the translator, which is responsible for translating $\boldsymbol{p}^S$ into target language prompt $\boldsymbol{p}^T$, thereby infusing $\boldsymbol{p}^T$ with multilingual knowledge. Specifically,
 the external corpus, comprising a source language dataset and a mixed target language dataset, serves as auxiliary training data. We select parallel sentences from this corpus, each containing a source and a random target language sentence, and concatenate them with the respective prompt $\boldsymbol{p}^S$ and $\boldsymbol{p}^T$. These concatenated pairs are then fed into the PLM, yielding two probabilities of predicted word across entire vocabulary. The training goal is to minimize KLD loss between these probabilities in the source and target languages. By concurrently learning from both tasks, MPT ultimately gets $\boldsymbol{p}^T$ via the optimized translator. This optimization allowing $\boldsymbol{p}^T$ to retain the capability of original task and multilingual ability, facilitating more precise inference in the target language during cross-lingual transfer.
 
 To verify the effectiveness of our proposed MPT, we implement empirical experiments in few-shot scenarios on XNLI, a benchmark for XLU. Additionally, we construct PAWS-15 as auxiliary data based on an external corpus PAWS-X. In the experimental results, MPT achieves 7.5\% and 3.1\% absolute performance gains under 4-shots and 64-shots settings, respectively. More remarkably, in comparison to vanilla prompting, MPT showcases an impressive 18.4\% enhancement in relative performance when transferring to languages considerably different from the source language.

 To summarize, our contributions are as follows:
 \begin{itemize}
    \item 
    We propose MPT, a novel and effective method for translating source language soft prompt into target language prompt. It contains multilingual knowledge and task knowledge to facilitate cross-lingual transfer.
    \item 
    We expand an external dataset into a parallel corpus in 15 languages and serve it as auxiliary training data in the multi-task training process for MPT.
    \item
    We conduct extensive experiments in different few-shot settings to compare and analyze MPT with existing methods. The results demonstrate the effectiveness and advancement of MPT in cross-lingual transfer.
\end{itemize}

\section{Related work}
\textbf{Cross-lingual transfer learning.}$\;$
The publication of multilingual PLMs, such as mBERT~\cite{devlin2018bert}, has catalyzed a growing interest in cross-lingual transfer learning~\cite{pfeiffer2020mad,lauscher2020zero}. This approach primarily focuses on leveraging knowledge learned from high-resource languages to better address challenges in low-resource languages. Wu et al.~\cite{Wu2019BetoBB} verify that the shared vocabulary of two languages remarkably promotes transferability. Qin et al.~\cite{ijcai2020p533} enforce model to align representations of different languages once by mixing contextual information. Chalkidis et al.~\cite{chalkidis-etal-2021} emphasize the importance of partially fine-tuning PLM parameters to avoid the catastrophic knowledge forgetting caused by full model fine-tuning. In contrast, with an external corpus, we adopt soft prompt with multilingual knowledge and task knowledge for better transfer.

\smallskip
\noindent
\textbf{Prompt learning.}$\;$
Since the advent of GPT-3~\cite{brown2020language}, prompt learning has received considerable attention, where the key is how to imitate the pre-training of PLMs and induce learned knowledge validly. Schick et al.~\cite{schick2020exploiting} design discrete prompt, which consists of interpretable natural language tokens. Several studies~\cite{lester2021power,li2021prefix} successively inject trainable vectors in PLMs to assist training. Building on these, Zhao et al.~\cite{zhao2021discrete} explore the use of prompting for enhancing cross-lingual transfer, where source language soft prompt is applied to all target languages. Qi et al.~\cite{qi2022enhancing} introduce cross-lingual hard prompt to encourage alignment of answer distributions across languages. In comparison, we exploit prompt translator to translate source language soft prompt into target prompt endowed with multilingual knowledge and task knowledge.

\section{Methodology}
\begin{figure*}[t]
 \centering
 \includegraphics[width=2.0\columnwidth]{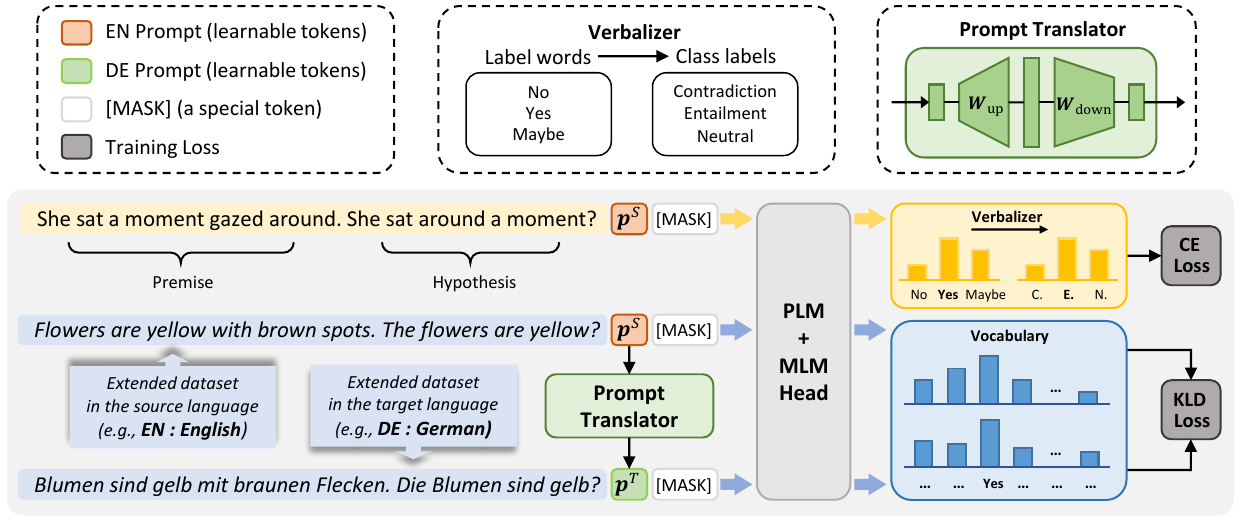}
 \caption{An overview of our proposed MPT, where a prompt translator is designed to translate source language prompt into target language prompt. In addition to original training data in yellow, we incorporate auxiliary data in blue from an extended dataset. MPT is optimized by minimizing the combined CE loss and KLD loss for classification task and cross-lingual alignment task.}
 \label{MPT}
\end{figure*}
The overall framework of the proposed MPT is depicted in Figure~\ref{MPT}, where a multilingual prompt translator is designed to translate source language prompt ($\boldsymbol{p}^S$) learned in source language (e.g., EN: English) to target language prompt ($\boldsymbol{p}^T$). To optimize the translator, our approach involves not just completing the original task using EN training data, but also leveraging an external corpus for an auxiliary cross-lingual alignment task. In this way, the refined $\boldsymbol{p}^T$ that we eventually obtain is imbued with original task knowledge and multilingual knowledge, thereby enhancing better inference on test data in target language (e.g., DE: German). The model is jointly optimized by minimizing the CE loss supervising the original task and the KLD loss regularizing the auxiliary task.

\subsection{Problem formulation}
This section delineates the original task process, namely cross-lingual Natural Language Inference (NLI) based on soft prompting. Formally speaking, let $\mathcal{M}$ and $\mathcal{V}$ represent multilingual PLM and its corresponding vocabulary, respectively. In NLI, every English training triple is denoted by $D = (X_P, X_H, Y)$, where $X_P$, $X_H$ and $Y$ refer to the premise sentence, the hypothesis sentence and the NLI category. For cross-lingual transfer, we take $S$ and $T$ to symbolize the source and target language separately. ${\mathcal{D}} = ({\mathcal{X}}_P, {\mathcal{X}}_H, \mathcal{Y})$ signifies training set in $S$ of XNLI. Following Schick et al.~\cite{schick-schutze-2021}, we reframe a typical NLI example $D$ as a cloze-style question $X$ by filling in a template with $X_P$ and $X_H$ and utilizing soft prompt as ``pseudo tokens" that are not part of $\mathcal{V}$:
\begin{small} 
\begin{eqnarray}
   X
   = \underline{X_P} \; . \; \underline{X_H} \; ? \; [\boldsymbol{p}] \; \underline{\hbox to 5mm{}} \; .\label{temp}
\end{eqnarray}
\end{small}
Here, $\mathcal{M}$ is used to fill in the blank ($\underline{\hbox to 3mm{}}$) occupied by a special token [MASK] in $X$. $\boldsymbol{p} \in {\mathbb{R}}^{m \times d} $ is a randomly initialized trainable soft prompt. $m$ is length of $\boldsymbol{p}$ and $d$ is hidden dimension of embedding layer in $\mathcal{M}$. We feed $\boldsymbol{p}$ forward through an LSTM and an MLP to obtain prompt we eventually use~\cite{liu2021gpt}. 

Formally speaking, we use $l$ as the size of $\mathcal{V}$ and $d$ also as the hidden dimension of the representations of a token. By feeding $X$ into $\mathcal{M}$, we can obtain the output $\boldsymbol{h}_X \in {\mathbb{R}}^d$, which is the contextualized representation of [MASK] in sentences. Further, the probability distribution of predicted word ${\rm{P}}^{\rm{MLM}}_X$ over $\mathcal{V}$ is calculated by MLM Head. To map ${\rm{P}}^{\rm{MLM}}_X$ to probability of label word ${\rm{P}}^{\rm{CLS}}_X$, we define a verbalizer $f: \mathcal{Z} \mapsto \mathcal{Y}$ as a mapping from the label word set $\mathcal{Z}$ to the NLI class label set $\mathcal{Y}$, where $\mathcal{Z}$ is comprised of a few words in $\mathcal{V}$. Utilizing ${\rm{P}}^{\rm{MLM}}_X$, we extract the probability corresponding to all label words in $\mathcal{Z}$ as the classification probability for $X$:
\begin{small} 
\begin{eqnarray}
\begin{aligned}
    &{\rm{P}}^{\rm{MLM}}_X 
    = \text{softmax}({{\mathbf{W_{lh}}}}\boldsymbol{h}_X) ,
    \\[1.3mm]
    &{\rm{P}}^{\rm{CLS}}_X 
    =  {\rm{P}}_{\mathcal{M}}([{\rm{MASK}}] = z|X) = {{\mathbf{W_{\mathcal{Z}}}}}{\rm{P}}^{\rm{MLM}}_X, z \in \mathcal{Z},
\end{aligned}
\end{eqnarray}
\end{small} 
where ${\mathbf{W_{lh}}} \in {\mathbb{R}}^{l \times d}$ symbolizes the parameters of Masked Language Model (MLM) Head. ${\mathbf{W_{\mathcal{Z}}}} \in {\mathbb{R}}^{n \times l}$ represents the parameter matrix for all label words in the output mapping layer, with $n$ indicating the size of $\mathcal{Z}$.

Given dataset $\mathcal{D}$ on XNLI, we perform the classification task to get ${\rm{P}}_{\mathcal{D}}^{\rm{CLS}}$, the probability of each label word. Based on ${\rm{P}}_{\mathcal{D}}^{\rm{CLS}}$, CE loss is calculated by the following equation:
\begin{small} 
\begin{eqnarray}
    {\mathcal{L}_{CE}} = -{g_{y}} \text{log}({\rm{P}}_{\mathcal{D}}^{\rm{CLS}}) ,
\end{eqnarray}
\end{small} 
where ${g_{y}}$ denotes the ``one-hot vector" of the gold labels $\mathcal{Y}$.

\subsection{Multilingual prompt translator}
In classification task, we acquire $\boldsymbol{p}^S$ learned in $S$, which possesses $S$ knowledge and classification task knowledge due to $S$ annotations of $\mathcal{D}$. However, directly using identical $\boldsymbol{p}^S$ in $T$ may not efficiently capture language-specific characteristics. As a result, we exploit a multilingual prompt translator designed to translate $\boldsymbol{p}^S$ into $\boldsymbol{p}^T$, reasonably transferring $S$ knowledge to $T$ knowledge and preserving task knowledge. For inference, $\boldsymbol{p}^T$ is utilized on $T$ test data. Formally, translator is parameterized with a two-layer perceptron as follows:
\begin{small} 
\begin{eqnarray}
\begin{aligned}
    \boldsymbol{p}^T 
    & = \textbf{Transl}(\boldsymbol{p}^S)
    \\[1.3mm]
    & = {\mathbf{W_{down}}}(\text{ReLU}({\mathbf{W_{up}}}{\boldsymbol{p}^S} + \rm{b_{up}})) + \rm{b_{down}} ,
\end{aligned}
\end{eqnarray}
\end{small} 
where ${\mathbf{W_{up}}} \in {\mathbb{R}}^{d \times 2d}$ and ${\mathbf{W_{down}}} \in {\mathbb{R}}^{2d \times d}$ are trainable matrices, $\rm{b_{up}} \in {\mathbb{R}}^{2d}$ and $\rm{b_{down}} \in {\mathbb{R}}^{d}$ are trainable bias terms, and $\text{ReLU($\cdot$)}$ is a non-linear activation function.

To optimize the translator, we perform a cross-lingual alignment task using an external corpus. It includes examples in $S$ and $T$, represented as ${\mathcal{\widetilde{D}}}^S = ({\mathcal{\widetilde{X}}}_P^S, {\mathcal{\widetilde{X}}}_H^S)$ and ${\mathcal{\widetilde{D}}}^T = ({\mathcal{\widetilde{X}}}_P^T, {\mathcal{\widetilde{X}}}_H^T)$, respectively. In a manner akin to the input construction in the classification task, the sentences from ${\mathcal{\widetilde{D}}}^S$ and ${\mathcal{\widetilde{D}}}^T$ are spliced with their corresponding prompt, $\boldsymbol{p}^S$ and $\boldsymbol{p}^T$. Therefore, the final constructed additional sentences in $S$ and $T$ as the second set of inputs are as follows:
\begin{small} 
\begin{eqnarray}
   {\widetilde{X}}^S
   = \underline{{\widetilde{X}}_P^S} \, . \, \underline{{\widetilde{X}}_H^S} \, ? \, [\boldsymbol{p}^S] \, \underline{\hbox to 3mm{}} \, .\quad
   {\widetilde{X}}^T
   = \underline{{\widetilde{X}}_P^T} \, . \, \underline{{\widetilde{X}}_H^T} \, ? \, [\boldsymbol{p}^T] \, \underline{\hbox to 3mm{}} \, .
\end{eqnarray}
\end{small} 

Similarly, we input ${\widetilde{X}}^S$ and ${\widetilde{X}}^T$  into PLM to generate hidden features $\boldsymbol{h}_{{\widetilde{X}}^S}$ and $\boldsymbol{h}_{{\widetilde{X}}^T}$ for the [MASK]. Afterwards, they are mapped back to vocabulary space through MLM head to get predicted word probability ${\rm{P}}^{\rm{MLM}}_{{\widetilde{X}}^S}$ and ${\rm{P}}^{\rm{MLM}}_{{\widetilde{X}}^T}$:
\begin{small} 
\begin{eqnarray}
\begin{aligned}
    &{\rm{P}}_{{\widetilde{X}}^S}^{\rm{MLM}} = \widetilde{\rm{P}}_{\mathcal{M}}^S([{\rm{MASK}}] = v|{\widetilde{X}}^S)= \text{softmax}({{\mathbf{W_{lh}}}}\boldsymbol{h}_{{\widetilde{X}}^S}), v \in \mathcal{V} ,
    \\[1.3mm]
    &{\rm{P}}_{{\widetilde{X}}^T}^{\rm{MLM}} =  \widetilde{\rm{P}}_{\mathcal{M}}^T([{\rm{MASK}}] = v|{\widetilde{X}}^T) = \text{softmax}({{\mathbf{W_{lh}}}}\boldsymbol{h}_{{\widetilde{X}}^T}), v \in \mathcal{V} .
\end{aligned}
\end{eqnarray}
\end{small} 

For ${\mathcal{\widetilde{D}}}^S$ and ${\mathcal{\widetilde{D}}}^T$ from the external corpus, we apply them for MLM to derive ${\rm{P}}_{{\mathcal{\widetilde{D}}}^S}^{\rm{MLM}}$ and ${\rm{P}}_{{\mathcal{\widetilde{D}}}^T}^{\rm{MLM}}$, which are the probability of each word in $\mathcal{V}$. We compute KLD loss for probability between source language and target language as follows:
\begin{small} 
\begin{eqnarray}
    {\mathcal{L}_{KLD}} = {\rm{KL}}({\rm{P}}_{{\mathcal{\widetilde{D}}}^S}^{\rm{MLM}}, {\rm{P}}_{{\mathcal{\widetilde{D}}}^T}^{\rm{MLM}}) + {\rm{KL}}({\rm{P}}_{{\mathcal{\widetilde{D}}}^T}^{\rm{MLM}}, {\rm{P}}_{{\mathcal{\widetilde{D}}}^S}^{\rm{MLM}}),
\end{eqnarray}
\end{small} 
where KL($\cdot$) symbolizes Kullback-Leibler divergence.

\subsection{Training and Inference}
\textbf{Training.}$\;$
We add ${\mathcal{L}_{CE}}$ in the classification task and ${\mathcal{L}_{KLD}}$ in the cross-lingual alignment task as the total loss, with a hyper-parameter $\alpha$ to balance the training process. The optimization of MPT is achieved by minimizing this total loss:
\begin{small} 
\begin{eqnarray}
    \mathcal{L}_{total} = \alpha \mathcal{L}_{CE} + (1 - \alpha)\mathcal{L}_{KLD} .
\end{eqnarray}
\end{small} 
\noindent
\textbf{Inference.}$\;$
In the inference, we apply the optimized $\boldsymbol{p}^T$ on $T$ test data. To enhance classification, $\boldsymbol{p}^T$ is concatenated after the input in target language. More concretely, to make the inference in DE examples, we firstly translate EN prompt into DE prompt via the translator, secondly, fill in the template with premises-hypothesis sentences followed by DE prompt, and lastly feed this combined input into PLMs to predict the relationship between the sentences in the framework of MPT.

\begin{table*}[ht]
\caption{Comparison results on XNLI under the few-shot setting in accuracy (\%).}
\label{tab:main result}
\scriptsize
\centering
\begin{tabular}{c|c|ccccccccccccccc|c}
\toprule
\rm{Shots} & \rm{Method} & \rm{AR} & \rm{BG} & \rm{DE} & \rm{EL} & \rm{EN} & \rm{ES} & \rm{FR} & \rm{HI} & \rm{RU} & \rm{SW} & \rm{TH} & \rm{TR} & \rm{UR} & \rm{VI} & \rm{ZH} & \rm{Avg.}\\
\midrule
\multirow{4}{*}{1} & \rm{FT} & 33.2 & 33.3 & 33.3 & 33.1 & 33.3 & 33.2 & 32.8 & 33.0 & 33.3 & 33.0 & 33.3 & 32.9 & 32.9 & 33.2 & 33.2 & 33.1 \\
& \rm{SP} & 35.4 & 36.4 & 36.4 & 36.6 & 36.5 & 37.6 & 37.9 & 36.0 & 37.5 & 34.1 & 35.9 & 34.7 & 35.0 & 35.5 & \textbf{36.7} & 36.1 \\
& \rm{PCT} & 33.2 & 35.4 & 34.8 & 35.1 & 35.9 & 35.3 & 35.7 & 34.6 & 36.2 & 33.8 & 34.6 & 34.3 & 33.1 & 34.9 & 35.0 & 34.8 \\
& \rm{MPT} & \textbf{37.0} & \textbf{38.5} & \textbf{37.8} & \textbf{38.1} & \textbf{38.6} & \textbf{38.1} & \textbf{38.7} & \textbf{37.2} & \textbf{38.5} & \textbf{36.5} & \textbf{37.1} & \textbf{37.6} & \textbf{37.3} & \textbf{37.9} & 35.7 & \textbf{37.6} \\
\midrule
\multirow{4}{*}{2} & \rm{FT} & 33.5 & 33.3 & 33.7 & 33.3 & 34.1 & 33.5 & 33.7 & 33.2 & 33.5 & 33.3 & 33.8 & 33.6 & 33.5 & 34.0 & 33.3 & 33.5 \\
& \rm{SP} & 36.6 & 37.9 & 38.0 & 38.2 & 38.0 & 38.0 & 38.3 & 36.2 & 38.9 & 34.3 & 37.5 & 34.6 & 35.2 & 37.2 & 36.7 & 37.0 \\
& \rm{PCT} & 34.1 & 39.0 & 39.1 & 38.2 & 39.9 & 40.6 & 40.5 & 37.9 & 39.9 & 36.5 & 37.2 & 36.9 & 34.7 & 37.9 & 37.1 & 38.0 \\
& \rm{MPT} & \textbf{41.6} & \textbf{42.8} & \textbf{40.8} & \textbf{43.2} & \textbf{43.2} & \textbf{42.5} & \textbf{42.8} & \textbf{40.4} & \textbf{43.3} & \textbf{36.8} & \textbf{40.5} & \textbf{41.0} & \textbf{41.1} & \textbf{41.4} & \textbf{38.2} & \textbf{41.3} \\
\midrule
\multirow{4}{*}{4} & \rm{FT} & 34.2 & 34.5 & 34.1 & 34.3 & 34.1 & 34.1 & 34.5 & 34.0 & 34.3 & 33.7 & 34.0 & 34.0 & 34.1 & 34.2 & 34.2 & 34.1  \\
& \rm{SP} & 37.4 & 39.7 & 39.2 & 39.7 & 40.2 & 38.9 & 40.5 & 37.1 & 40.6 & 35.3 & 38.1 & 35.3 & 36.9 & 37.2 & 38.9 & 38.3  \\
& \rm{PCT} & 33.9 & 37.2 & 37.0 & 36.2 & 37.0 & 37.7 & 37.5 & 36.4 & 37.4 & 34.2 & 34.7 & 34.7 & 33.5 & 35.0 & 35.6 & 35.9  \\
& \rm{MPT} & \textbf{42.9} & \textbf{43.6} & \textbf{44.3} & \textbf{43.6} & \textbf{45.5} & \textbf{44.2} & \textbf{44.1} & \textbf{42.8} & \textbf{44.1} & \textbf{40.2} & \textbf{43.4} & \textbf{42.7} & \textbf{42.4} & \textbf{43.8} & \textbf{43.1} & \textbf{43.4}  \\
\midrule
\multirow{4}{*}{8} & \rm{FT} & 32.8 & 32.7 & 32.8 & 32.9 & 32.7 & 32.6 & 33.0 & 33.3 & 32.7 & 33.0 & 33.2 & 33.0 & 33.1 & 32.5 & 32.4 & 32.8  \\
& \rm{SP} & 37.4 & 39.6 & 38.1 & 39.1 & 40.0 & 38.8 & 39.2 & 36.5 & 40.3 & 35.6 & 38.5 & 35.3 & 36.5 & 37.8 & 37.1 & 38.0  \\
& \rm{PCT} & 40.2 & 40.6 & 40.9 & 41.7 & 41.9 & 41.7 & 41.6 & 41.0 & 40.6 & 39.2 & 41.4 & \textbf{41.4} & 38.4 & 41.3 & \textbf{41.2} & 40.9  \\
& \rm{MPT} & \textbf{42.7} & \textbf{43.0} & \textbf{41.9} & \textbf{42.4} & \textbf{43.1} & \textbf{42.3} & \textbf{42.1} & \textbf{40.8} & \textbf{42.6} & \textbf{39.4} & \textbf{41.9} & 40.1 & \textbf{40.7} & \textbf{42.2} & 40.2 & \textbf{41.7}  \\
\midrule
\multirow{4}{*}{16} & \rm{FT} & 33.6 & 33.4 & 33.3 & 33.5 & 34.1 & 33.4 & 33.3 & 33.4 & 33.4 & 33.6 & 33.6 & 33.4 & 33.5 & 33.3 & 33.5 & 33.5  \\
& \rm{SP} & 39.5 & 39.9 & 39.1 & 40.4 & 41.1 & 40.2 & 40.4 & 37.4 & 40.7 & 37.1 & 39.3 & 36.5 & 36.0 & 38.2 & 38.3 & 38.9  \\
& \rm{PCT} & \textbf{43.6} & 40.8 & 36.9 & \textbf{45.7} & \textbf{46.5} & 41.5 & 44.3 & \textbf{44.8} & 42.4 & 40.1 & \textbf{43.9} & \textbf{43.7} & \textbf{42.5} & \textbf{44.7} & \textbf{44.8} & 43.1  \\
& \rm{MPT} & 43.5 & \textbf{43.8} & \textbf{44.0} & 43.9 & 45.2 & \textbf{44.2} & \textbf{44.3} & 42.9 & \textbf{43.4} & \textbf{40.2} & 42.5 & 41.8 & 42.0 & 43.4 & 42.2 & \textbf{43.1}  \\
\midrule
\multirow{4}{*}{32} & \rm{FT} & 36.1 & 36.3 & 35.7 & 35.7 & 36.5 & 36.2 & 36.0 & 35.5 & 35.9 & 35.0 & 35.6 & 36.0 & 35.4 & 36.1 & 36.3 & 35.9  \\
& \rm{SP} & 41.7 & 43.4 & 42.8 & 42.3 & 44.9 & 42.9 & 43.3 & 39.2 & 43.5 & 37.7 & 40.2 & 41.1 & 39.8 & 43.0 & 39.8 & 41.7  \\
& \rm{PCT} & 45.7 & 45.4 & 44.4 & \textbf{47.4} & 49.6 & 45.5 & \textbf{48.8} & \textbf{46.7} & 45.5 & 40.3 & 41.6 & 44.3 & 42.9 & 46.7 & 45.6 & 45.4  \\
& \rm{MPT} & \textbf{47.1} & \textbf{47.6} & \textbf{47.9} & 47.1 & \textbf{49.9} & \textbf{49.1} & 48.2 & 46.3 & \textbf{47.3} & \textbf{43.3} & \textbf{47.2} & \textbf{47.2} & \textbf{45.3} & \textbf{49.0} & \textbf{47.1} & \textbf{47.3}  \\
\midrule
\multirow{4}{*}{64} & \rm{FT} & 41.4 & 41.2 & 41.5 & 40.7 & 42.6 & 41.4 & 40.8 & 41.2 & 40.2 & 40.6 & 40.7 & 41.4 & 40.5 & 41.7 & 41.0 & 41.1  \\
& \rm{SP} & 43.9 & 44.2 & 47.5 & 45.1 & 50.5 & 47.9 & 48.6 & 41.8 & 43.7 & 41.3 & 45.9 & 45.3 & 42.6 & 47.6 & 45.1 & 45.4  \\
& \rm{PCT} & 48.1 & 50.2 & 49.3 & 50.6 & 51.1 & 50.9 & 51.3 & 47.6 & 49.1 & 44.6 & 47.3 & 47.4 & 44.0 & 49.7 & 48.2 & 48.6  \\
& \rm{MPT} & \textbf{50.7} & \textbf{52.7} & \textbf{53.1} & \textbf{52.2} & \textbf{55.4} & \textbf{53.8} & \textbf{53.1} & \textbf{50.2} & \textbf{51.0} & \textbf{46.2} & \textbf{51.5} & \textbf{50.4} & \textbf{49.1} & \textbf{53.0} & \textbf{52.3} & \textbf{51.7}  \\
\midrule
\multirow{4}{*}{128} & \rm{FT} & 43.9 & 44.4 & 44.4 & 43.7 & 46.3 & 44.6 & 44.5 & 42.9 & 42.7 & 41.7 & 43.0 & 43.2 & 42.7 & 44.9 & 43.8 & 43.8  \\
& \rm{SP} & 46.2 & 46.8 & 47.8 & 47.6 & 53.0 & 48.5 & 49.6 & 47.3 & 45.5 & 41.7 & 47.5 & 46.4 & 44.5 & 45.6 & 48.7 & 47.1  \\
& \rm{PCT} & 50.4 & 51.9 & 52.8 & 53.4 & 55.0 & 53.8 & 53.3 & 51.5 & 51.7 & 47.0 & 50.0 & 50.9 & 47.9 & 51.7 & 51.2 & 51.5  \\
& \rm{MPT} & \textbf{53.2} & \textbf{56.1} & \textbf{56.0} & \textbf{55.4} & \textbf{57.4} & \textbf{56.4} & \textbf{56.6} & \textbf{53.5} & \textbf{54.8} & \textbf{48.6} & \textbf{54.0} & \textbf{53.1} & \textbf{51.8} & \textbf{55.2} & \textbf{55.4} & \textbf{54.5}  \\
\midrule
\multirow{4}{*}{256} & \rm{FT} & 53.3 & 55.6 & 56.5 & 55.0 & 58.8 & 56.9 & 56.4 & 52.5 & 53.6 & 50.5 & 52.6 & 53.8 & 51.3 & 55.0 & 53.0 & 54.3  \\
& \rm{SP} & 52.7 & 55.2 & 49.6 & 53.7 & 59.5 & 55.0 & 55.3 & 50.6 & 51.4 & 46.5 & 53.4 & 46.1 & 44.9 & 52.8 & 51.5 & 51.9  \\
& \rm{PCT} & 54.7 & 56.7 & 56.3 & 57.9 & 60.3 & 58.3 & 58.3 & 54.6 & 55.2 & \textbf{51.6} & 55.6 & 54.6 & 52.6 & 57.4 & 55.8 & 56.0  \\
& \rm{MPT} & \textbf{59.0} & \textbf{61.1} & \textbf{60.9} & \textbf{60.6} & \textbf{65.8} & \textbf{63.0} & \textbf{61.9} & \textbf{57.6} & \textbf{60.6} & 50.7 & \textbf{59.2} & \textbf{57.8} & \textbf{56.1} & \textbf{60.7} & \textbf{60.8} & \textbf{59.7}  \\
\bottomrule
\end{tabular}
\end{table*}

\begin{figure*}[htbp]
 \centering
 \begin{minipage}[t]{0.66\columnwidth}
 \centering
 \includegraphics[width=\columnwidth]{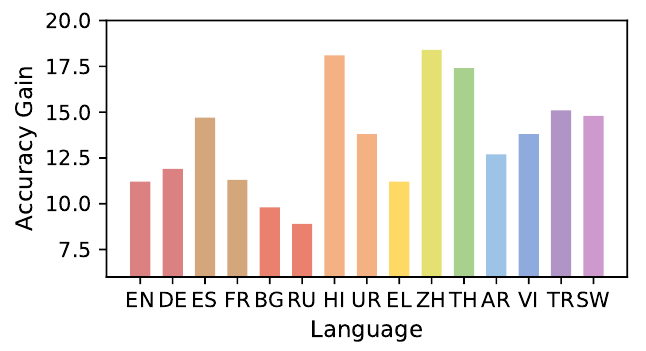}
 \caption{Accuracy gain of MPT relative to SP among different languages (\%).}
 \label{data_lang}
 \end{minipage}
 \hspace{10pt}
 \begin{minipage}[t]{1.20\columnwidth}
 \centering
 \includegraphics[width=\columnwidth]{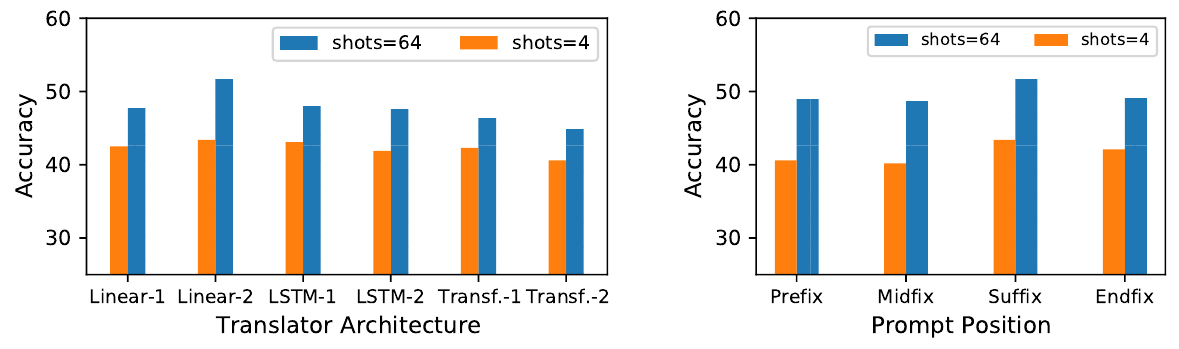}
 \caption{Ablations of the architecture of multilingual prompt translator and the position of soft prompt on MPT performance in accuracy (\%).}
 \label{abla_1}
 \end{minipage}
\end{figure*}

\section{Experiments}
\subsection{Dataset}
We evaluate our approach on XNLI~\cite{conneau2018xnli}, an XLU benchmark covering 15 languages. Each example consists of a premise-hypothesis sentence pair and a label showing their relationship, classified as entailment, contradiction, or neutral. To simulate low-resource scenarios, we follow Zhao et al.~\cite{zhao2021discrete} and create new training and development sets by randomly sampling $k \in \{ 1, 2, 4, 8, 16, 32, 64, 128, 256 \}$ instances from each label category, 3$k$ instances in total, whereas test set remains unchanged. In cross-lingual transfer, the source language is EN used as the training and development sets. The remaining 14 languages serve as target languages, which are used in the test set with EN. Besides, we extend PAWS-X~\cite{yangetal2019} to construct a dataset called PAWS-15 to serve as external parallel corpus. To be specific, we randomly select 500 English examples from training set in PAWS-X and translate them into 14 randomly selected target language sentences via Google Translate. Consequently, within PAWS-15, there are 1,000 samples, comprising 500 in English and 500 in mixed-language, which correspond to each other as parallel sentences. See Appendix 1.1 for details on XNLI and PAWS-15.

\subsection{Setup}
We adopt pre-trained XLM-RoBERTa-base~\cite{conneau2019unsupervised} in Huggingface Transformers as the backbone model. For MPT, we set the loss weight to 0.5, the prompt length to 4 and the maximum sequence length to 256. All models are trained for 50 epochs with a batch-size of 24 and a learning rate of 4e-5 via AdamW optimizer on a single GeForce GTX 3090Ti GPU. Additionally, we run the model with 5 random seeds ($\{1, 2, 3, 4, 5\}$) and report the average results due to the large variance when evaluating few-shot cross-lingual transfer. Appendix 1.2 shows experimental implementation details.

\subsection{Comparison with Existing Methods}
\label{results}
We assess MPT and compare it with the following baselines which employ prompt learning for cross-lingual transfer, including Fine-Tuning (FT), Soft Prompting (SP)~\cite{zhao2021discrete} and Prompt-learning from Cross-lingual Templates (PCT)~\cite{qi2022enhancing}.

Extensive experiments are conducted under few-shot settings on XNLI. The results are exhibited in Table~\ref{tab:main result} and concrete results are in Appendix 1.3. As seen from average performance, MPT consistently surpasses all baselines for most data scales. Notably, MPT outperforms robust PCT by an average of 4.3\% even with limited data ($k = { 1, 2, 4}$). We conjecture that the considerable improvement is due to an auxiliary external parallel corpus comprising 500 pairs of parallel data. This greatly enhances the alignment of cross-lingual probabilities, which in turn facilitates effective learning of multilingual knowledge. In addition, when $ k = \{ 32 \sim 256\} $, MPT confers great boosts with \{1.9\%, 3.1\%, 3.0\%, 3.7\%\} higher accuracy compared with PCT, respectively.

Next, we analyze MPT from the perspective of language. As Figure~\ref{data_lang} shows, we categorize 15 languages into 11 families, each represented by distinct colors. For instance, EN and DE are part of Germanic, whereas ZH belongs to Chinese. We compare the accuracy gains of MPT over SP under 32-shots for different families. Notably, MPT consistently outperforms SP across all languages, with a substantial 18.4\% increase in relative performance for ZH. This trend is particularly evident in ZH, TH, HI, and TR, where the gains exhibited by MPT are higher compared to EN. We attribute this phenomenon to an intuitive factor that these languages differ significantly from EN, leading to a performance decline when tested directly in SP. However, our cross-lingual alignment task facilitates target prompt learning of multilingual knowledge resulting in greater improvement. Furthermore, SP underperforms in low-resource languages i.e. SW and UR, as they are not well-trained during pre-training of PLM. In contrast, MPT displays remarkable improvements with relative gains of 14.8\% and 13.8\% in SW and UR respectively. This proves MPT effectively compensates for this drawback with the multilingual knowledge learned through the target prompt.

\subsection{Ablation Study}

\begin{figure*}[htbp]
 \centering
 \includegraphics[width=2.0\columnwidth]{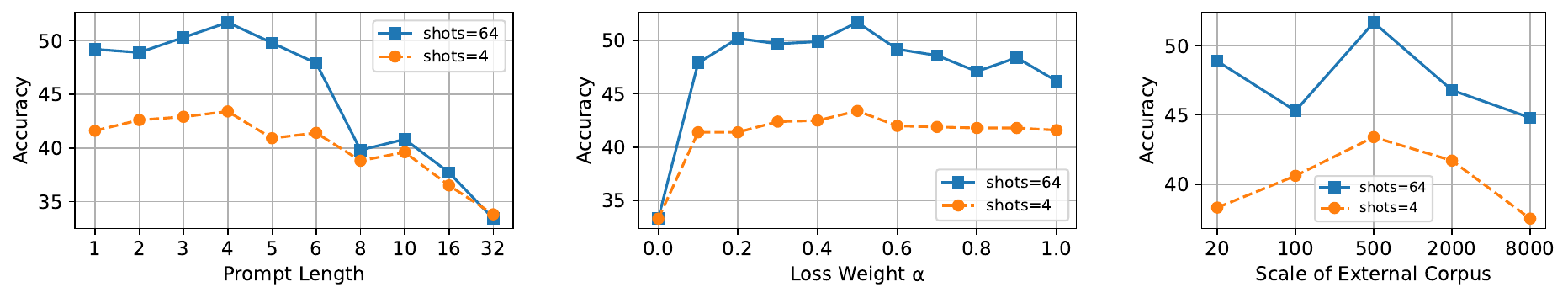}
 \caption{Ablations of various hyper-parameters on MPT performance measured by accuracy (\%). From left to right, we sequentially illustrate the impact of changes in prompt length, loss weight $\alpha$, and external data size on MPT performance.}
 \label{abla_2}
\end{figure*}

\noindent
\textbf{Translator architecture.}$\;$
In order to verify the effectiveness of translator and the performance of different architectures as translators, we design three network structures: Linear, LSTM, and Transformer. For each of them, we try both one-layer (*-1) and two-layer (*-2) configurations. Figure~\ref{abla_1} shows Linear-2 works best, which just confirms the conclusion proposed by Conneau et al.~\cite{conneau-etal-2020} that language representation across languages of PLM only needs to map to each other via simple linear layer. Secondly, Linear-2 avoids overfitting under the few-shot scenario. However, Linear-1 is too naive to learn multilingual knowledge and task knowledge, leading to slightly underperforming.
\smallskip

\noindent
\textbf{Prompt position.}$\;$
According to Equation~\ref{temp}, we investigate the impact of 4 different positions in Equation~\ref{temp8} on MPT. As seen in Figure~\ref{abla_1}, placing prompt at $\textsf{Suffix}$ yields outstanding results. Intuitively, the prompt placed between the sentence pair and the blank tends to understand contextualized semantics and guide PLM to fill in the blank reasonably.
\begin{small} 
\begin{eqnarray}
    X=\textsf{Prefix} \; \underline{X_P} \,.\; \textsf{Midfix} \; \underline{X_H}\,? \;\textsf{Suffix} \; \underline{\hbox to 5mm{}} \; \textsf{Endfix}\,.\label{temp8}
\end{eqnarray}
\end{small}

\noindent
\textbf{Prompt length.}$\;$
We train MPT while varying prompt lengths. Figure~\ref{abla_2} reveals that a length of 4 is the most appropriate. As length increases, trainable parameters increase but examples are so few that MPT overfits causing a dramatic performance decline. Conversely, with too little prompt, learned knowledge is not enough to exhibit satisfactory outcomes.
\smallskip

\noindent
\textbf{Loss weight.}$\;$
The influence of the cross-lingual alignment task is explored by sequentially changing $\alpha$ from a range [0.0, 1.0]. Figure~\ref{abla_2} illustrates performance is optimal when $\alpha$ = 0.5, indicating the alignment task and the original classification task are equally important and complementary.
\smallskip

\noindent
\textbf{Scale of external corpus.}$\;$
Further experiments are conducted to explore impact of external corpus. To keep training time consistent, we perform 15 and 4 epochs for 2000 and 8000 data scales, while 50 epochs for the rest. Figure~\ref{abla_2} shows that MPT becomes more productive with more additional data. Nevertheless, accuracy starts to diminish when scale increases to a certain level. Hence, we only need a few handy data to achieve desired results, which is a controllable cost.
\begin{table*}[htbp]
\caption{Intrinsic analysis of prime components of MPT under 64-shots. The p-value is calculated by two-tailed t-tests.}
\label{tab:mpt_ablation_64}
\setlength{\tabcolsep}{5pt}
\scriptsize
\centering
\begin{tabular}{l|cccccccccccccccc|c}
\toprule
Models                & AR   & BG   & DE   & EL   & EN   & ES   & FR   & HI   & RU   & SW   & TH   & TR   & UR   & VI   & ZH   & Avg. & p-value    \\
\midrule
Original MPT          & \textbf{50.7} & \textbf{52.}7 & \textbf{53.1} & \textbf{52.2} & \textbf{55.4} & \textbf{53.8} & \textbf{53.1} & \textbf{50.2} & \textbf{51.0} & \textbf{46.2} & \textbf{51.5} & \textbf{50.4} & \textbf{49.1} & \textbf{53.0} & \textbf{52.3} & \textbf{51.7} & -          \\
w/o consistency loss  & 45.8 & 46.3 & 44.8 & 47.7 & 51.0 & 48.4 & 47.7 & 47.5 & 48.2 & 38.1 & 46.8 & 44.4 & 42.8 & 45.8 & 47.6 & 46.2 & 9.7e-7   \\
w/o external data     & 47.3 & 48.7 & 49.4 & 48.8 & 52.8 & 50.3 & 50.6 & 46.8 & 47.1 & 42.2 & 48.6 & 47.2 & 46.0 & 49.4 & 49.5 & 48.3 & 2.0e-4 \\
w/o prompt translator & 48.1 & 48.4 & 51.1 & 48.5 & 53.5 & 51.1 & 51.0 & 46.4 & 46.1 & 42.9 & 49.7 & 47.8 & 45.7 & 50.7 & 50.4 & 48.8 & 1.8e-3 \\
w/o MPT framework      & 43.9 & 44.2 & 47.5 & 45.1 & 50.5 & 47.9 & 48.6 & 41.8 & 43.7 & 41.3 & 45.9 & 45.3 & 42.6 & 47.6 & 45.1 & 45.4 & 1.9e-8   \\
\bottomrule
\end{tabular}
\end{table*}
\smallskip

\noindent
\textbf{Intrinsic analysis.}$\;$
To explore the role of prime components in MPT, an intrinsic analysis is further carried out and ablation results under 64-shots are presented in Table ~\ref{tab:mpt_ablation_64}. The results under 4-shots are reported in Appendix 1.3. When consistency loss is omitted, i.e. $\alpha$ = 1, our method experiences a performance drop of 5.5\%, indicating that our cross-lingual alignment task is beneficial. When external data is discarded and replaced with XNLI data without annotations, the method appears inferior to MPT owing to the lack of external data that brings data diversity. When translator is removed, the average accuracy drops by 2.9\%. We estimate that this is because target prompt makes it hard to learn task knowledge without translator. When MPT framework is deleted, the results show a dramatic 6.3\% reduction, which implies that it is difficult for target prompt to learn multilingual knowledge in this case.

\section{Conclusion}
In this work, we propose MPT, a novel and effective prompting approach for XLU. MPT promisingly fosters cross-lingual transfer via a multilingual prompt translator, which aims to translate source language soft prompt into target language soft prompt. We adopt a multi-task strategy based on an external corpus to optimize MPT, thus validly retaining task knowledge and converting language knowledge in prompt. In extensive experiments on few-shot XNLI, MPT achieves encouraging improvements compared with previous approaches, which demonstrates the effectiveness of our method.

\bibliographystyle{IEEEbib}
\bibliography{icme2023template}

\begin{thebibliography}{10}

\bibitem{devlin2018bert}
J.~Devlin, M.-W. Chang, K.~Lee, et~al.,
\newblock ``{BERT}: Pre-training of deep bidirectional transformers for language understanding,''
\newblock in {\em NAACL}, 2019, pp. 4171--4186.

\bibitem{brown2020language}
T.~Brown, B.~Mann, N.~Ryder, et~al.,
\newblock ``Language models are few-shot learners,''
\newblock in {\em NeurIPS}, 2020, pp. 1877--1901.

\bibitem{conneau2019cross}
A.~Conneau and G.~Lample,
\newblock ``Cross-lingual language model pretraining,''
\newblock in {\em NeurIPS}, 2019, pp. 7059--7069.

\bibitem{conneau2019unsupervised}
A.~Conneau, K.~Khandelwal, N.~Goyal, et~al.,
\newblock ``Unsupervised cross-lingual representation learning at scale,''
\newblock in {\em ACL}, 2020, pp. 8440--8451.

\bibitem{zhao2021discrete}
M.~Zhao and H.~Sch{\"u}tze,
\newblock ``Discrete and soft prompting for multilingual models,''
\newblock in {\em EMNLP}, 2021, pp. 8547--8555.

\bibitem{li2022learning}
J.~Li, T.~Tang, J.~Nie, et~al.,
\newblock ``Learning to transfer prompts for text generation,''
\newblock {\em arXiv preprint arXiv:2205.01543}, 2022.

\bibitem{fu2022polyglot}
J.~Fu, S.~Ng, and P.~Liu,
\newblock ``Polyglot prompt: Multilingual multitask promptraining,''
\newblock {\em arXiv preprint arXiv:2204.14264}, 2022.

\bibitem{qi2022enhancing}
K.~Qi, H.~Wan, J.~Du, et~al.,
\newblock ``Enhancing cross-lingual natural language inference by prompt-learning from cross-lingual templates,''
\newblock in {\em ACL}, 2022, pp. 1910--1923.

\bibitem{liu2021gpt}
X.~Liu, Y.~Zheng, Z.~Du, et~al.,
\newblock ``{GPT} understands, too,''
\newblock {\em arXiv preprint arXiv:2103.10385}, 2021.

\bibitem{pfeiffer2020mad}
J.~Pfeiffer, I.~Vuli{\'c}, I.~Gurevych, et~al.,
\newblock ``{MAD-X}: An adapter-based framework for multi-task cross-lingual transfer,''
\newblock in {\em EMNLP}, 2020, pp. 7654--7673.

\bibitem{lauscher2020zero}
A.~Lauscher, V.~Ravishankar, I.~Vuli{\'c}, et~al.,
\newblock ``From zero to hero: {O}n the limitations of zero-shot language transfer with multilingual {T}ransformers,''
\newblock in {\em EMNLP}, 2020, pp. 4483--4499.

\bibitem{Wu2019BetoBB}
S.~Wu and M.~Dredze,
\newblock ``{B}eto, {B}entz, {B}ecas: The surprising cross-lingual effectiveness of {BERT},''
\newblock in {\em EMNLP-IJCNLP}, 2019, pp. 833--844.

\bibitem{ijcai2020p533}
L.~Qin, M.~Ni, Y.~Zhang, et~al.,
\newblock ``{C}o{SDA}-{ML}: Multi-lingual code-switching data augmentation for zero-shot cross-lingual {NLP},''
\newblock in {\em IJCAI}, 2020, pp. 3853--3860.

\bibitem{chalkidis-etal-2021}
I.~Chalkidis, M.~Fergadiotis, and I.~Androutsopoulos,
\newblock ``{M}ulti{EURLEX} - {A} multi-lingual and multi-label legal document classification dataset for zero-shot cross-lingual transfer,''
\newblock in {\em EMNLP}, 2021, pp. 6974--6996.

\bibitem{schick2020exploiting}
T.~Schick and H.~Sch{\"u}tze,
\newblock ``Exploiting cloze-questions for few-shot text classification and natural language inference,''
\newblock in {\em ACL}, 2021, pp. 255--269.

\bibitem{lester2021power}
B.~Lester, R.~Al-Rfou, and N.~Constant,
\newblock ``The power of scale for parameter-efficient prompt tuning,''
\newblock in {\em EMNLP}, 2021, pp. 3045--3059.

\bibitem{li2021prefix}
X.~L. Li and P.~Liang,
\newblock ``{P}refix-{T}uning: Optimizing continuous prompts for generation,''
\newblock in {\em EMNLP-IJCNLP}, 2021, pp. 4582--4597.

\bibitem{schick-schutze-2021}
T.~Schick and H.~Sch{\"u}tze,
\newblock ``It{'}s not just size that matters: Small language models are also few-shot learners,''
\newblock in {\em NAACL}, 2021, pp. 2339--2352.

\bibitem{conneau2018xnli}
A.~Conneau, R.~Rinott, G.~Lample, et~al.,
\newblock ``{XNLI}: Evaluating cross-lingual sentence representations,''
\newblock in {\em EMNLP}, 2018, pp. 2475--2485.

\bibitem{yangetal2019}
Y.~Yang, Y.~Zhang, C.~Tar, et~al.,
\newblock ``{PAWS}-{X}: A cross-lingual adversarial dataset for paraphrase identification,''
\newblock in {\em EMNLP-IJCNLP}, 2019, pp. 3687--3692.

\bibitem{conneau-etal-2020}
A.~Conneau, S.~Wu, H.~Li, et~al.,
\newblock ``Emerging cross-lingual structure in pretrained language models,''
\newblock in {\em ACL}, 2020, pp. 6022--6034.

\end{thebibliography}

\end{document}